\documentclass[11pt]{article}

\usepackage[preprint]{acl}

\usepackage{times}
\usepackage{latexsym}
\usepackage{enumitem}
\usepackage[T1]{fontenc}

\usepackage[utf8]{inputenc}

\usepackage{microtype}

\usepackage{inconsolata}
\usepackage{algorithm}
\usepackage{algpseudocode}
\usepackage{graphicx}
\usepackage{amsmath}
\usepackage{amssymb}
\usepackage{multirow}

\usepackage[most]{tcolorbox}
\usepackage{xcolor}
\usepackage{booktabs}
\usepackage{tabularx}
\usepackage{xcolor}
\usepackage{array}

\newcolumntype{Y}{>{\raggedright\arraybackslash}X}
\title{Reducing Hallucinated Transcripts in Whisper via Hallucination Space Projection}

\author{Maryam Abbasihafshejani \\
  The university of Texas at San Antonio  \\\And
 Murtuza Jadliwala, \\
  The university of Texas at San Antonio }

\begin{document}
\maketitle
\begin{abstract}
    Whisper is a widely used foundation model for Automatic Speech Recognition (ASR), providing robust transcription across languages and acoustic conditions. However, its generative decoder can produce fluent but hallucinated transcripts for inputs with little or no speech, which may be mistakenly treated as valid by downstream systems.

We propose a training-free, inference-time method to reduce Whisper hallucinations using low-rank decoder activation projection. The method estimates a compact hallucination-associated subspace from decoder activations on non-speech calibration data and, during inference, projects hidden states away from this subspace. The learned subspace is fixed and evaluated across multiple speech and non-speech benchmarks.

The projection increases Whisper’s no-speech probability on non-speech inputs, making hallucinated outputs easier to reject while exposing the HR--WER--FRR trade-off. We evaluate two variants: \textit{always-on}, which applies projection to all inputs, and \textit{gated}, which applies it only when Whisper predicts an input is likely non-speech. The always-on variant reduces average hallucination rate (HR) from 31.31\% to 2.44\%, while the gated variant reduces HR to 3.74\% with lower false rejection of real speech. These correspond to relative HR reductions of 92.21\% and 88.05\%, respectively.
We use LibriSpeech as a standard speech benchmark to measure whether hallucination suppression degrades recognition of genuine speech. Across model/split settings, gated projection incurs absolute word error rate (WER) increases of 0.33--4.39 percentage points and false-rejection rate (FRR) of 0.41--9.97\%, exposing a controllable HR--WER--FRR trade-off. However, in deployment, uncertain segments can be retained and flagged for re-transcription, verification, or human review rather than permanently discarded, thereby reducing the practical cost of false rejection.

\end{abstract}

\section{Introduction}

ASR systems are designed to convert speech into text, but many audio segments (e.g., silence, music, environmental sounds, and background noise) contain no speech at all and thus should not have any corresponding text transcript. A reliable ASR system should, therefore, produce accurate text when speech is present in the input audio and remain empty when it is not.

This requirement is especially critical for \emph{Whisper}~\citep{radford2023robust}, a state-of-the-art and widely used foundation model for ASR and robust speech processing. Whisper uses a generative sequence-to-sequence decoder, which gives it strong transcription ability across diverse conditions, but also allows it to sometimes emit fluent text even when the input contains no speech. For non-speech audio, where the correct transcript should be ``empty'', any generated text can thus be considered hallucination by the model.
This transcription failure scenario is especially problematic because the output often appears plausible. Whisper may generate acknowledgments, subtitle-style endings, conversational fillers, or repeated fragments that can be mistaken for real speech. Koenecke et al.~\citep{koenecke2024careless} show that these hallucinations are systematic and depend on both the acoustic properties of the input audio and the decoding behavior of the ASR model.

Whisper performs transcript generation and no-speech prediction within the same auto-regressive decoder, rather than relying on a separately trained speech activity detector. During decoding, Whisper assigns probabilities to a special no-speech token, commonly reported as \textit{no\_speech\_prob}. This score is used by Whisper's decoding-time rejection heuristic to suppress transcription when the input is likely to contain no speech. As a result, the decoder both generates text and provides the evidence used to decide whether an audio segment should be rejected as non-speech~\citep{radford2023robust}.

\emph{In this paper, our focus is to investigate whether non-speech hallucinations by Whisper can be reduced by editing its decoder activations during inference}. Our proposed method estimates a low-rank hallucination-associated direction from non-speech calibration examples and projects intermediate decoder hidden states away from this direction. The projection basis is estimated once and kept fixed, requiring no parameter updates. Our comprehensive evaluation experiments using the ESC-50, UrbanSound8K, FSD50K, and LibriSpeech audio benchmark 
datasets have shown significant improvement in the accurate rejection of non-speech audio samples, with limited impact on the speech transcription capability of the model.
We make the following contributions in this paper:
\begin{enumerate}[noitemsep]
    \item We introduce a training-free decoder activation projection method that reduces Whisper hallucinations on non-speech audio by estimating a low-rank calibration subspace and applying it at inference time without parameter updates.
    \item We evaluate on non-speech and speech benchmarks, showing that the compact decoder projection reduces hallucinated transcripts while preserving ASR performance and outperforming baselines without external Voice Activity Detection (VAD), phrase filtering, or fine-tuning.
\end{enumerate}

    
    


\section{Background and Related Works}
\label{sec:related_work}

\subsection{Whisper as a Foundation Model for ASR}
Whisper~\citep{radford2023robust} is a generative ASR model trained on 680K hours of weakly supervised multilingual and multitask audio data. It achieves strong zero-shot performance across languages, domains, and acoustic conditions, making it a widely used foundation model for robust speech transcription.
Whisper uses a Transformer encoder-decoder architecture. The encoder maps audio to acoustic representations, while the auto-regressive decoder generates text conditioned on the audio and previous tokens. This unified formulation combines transcription and no-speech estimation within the same auto-regressive decoder. 
However, because decoding is generative, the model can still assign high probability to fluent token sequences when acoustic evidence for speech is weak or absent, rather than suppressing generation through the no-speech decision. This failure mode produces hallucinated transcripts on non-speech audio and motivates decoder-level mitigation.
\subsection{Whisper Hallucinations on Non-Speech Audio}

Although Whisper performs well on speech inputs, recent studies show that it can generate fluent transcripts even when no speech is present~\citep{baranski2025investigation}. This differs from standard ASR errors measured by WER, which compares a prediction to a non-empty reference transcript. For non-speech audio, the correct transcript is empty, so any emitted text is hallucinated.

Prior work suggests that these hallucinations are systematic rather than random. Koenecke et al.~\citep{koenecke2024careless} report that about 1\% of Whisper transcriptions contain unsupported content, with roughly 38\% of such cases involving potentially harmful or misleading statements. They also observe higher hallucination rates in recordings with long pauses or weak vocal activity. In a non-speech setting, Barański et al.~\citep{baranski2025investigation} report that Whisper large-v3 produces non-empty transcriptions for 40.3\% of inputs, often as repeated acknowledgments, subtitle-style phrases, fillers, applause markers, or animal-sound onomatopoeia. They further show that hallucination rates vary with input duration, indicating a role for auto-regressive decoding dynamics.

These findings motivate our focus on non-speech hallucinations. Since the reference transcript is empty, hallucination can be measured directly by whether Whisper emits any text.

\subsection{Existing Mitigation Strategies}

Existing Whisper hallucination mitigation methods intervene at the input, output, model, or decoding level. VAD-based systems such as WhisperX~\citep{bain2023whisperx} remove non-speech regions before decoding, but depend on an external segmentation model. Post-processing methods, including the \emph{Bag of Hallucinations (BoH)}~\citep{baranski2025investigation}, filter hallucinated phrases after decoding, but may miss unseen hallucinations or remove valid speech.

Other approaches modify Whisper more directly. Fine-tuning methods such as Calm-Whisper~\citep{wang2025calm} update hallucination-related decoder components using non-speech audio with empty transcripts, but require additional training and may overfit to the fine-tuning data. Decoding heuristics based on no-speech probability, log probability, or compression ratio~\citep{radford2023robust} reject uncertain outputs, but operate only on output-level scores. In contrast, our method intervenes on decoder activations at inference time without external VAD, phrase filtering, or parameter updates.

Recent work on language and vision-language models shows that hidden-state interventions can reduce hallucinations by suppressing representations associated with unsupported generation~\citep{yang2025nullu,dastmalchi2026fighting}. Whisper non-speech hallucination differs because the same autoregressive decoder controls both transcript generation and no-speech prediction. We therefore apply a decoder-level projection conditioned on Whisper's own no-speech estimates during decoding.
\section{Proposed Method: Gated Low-Rank Decoder Projection}
We propose a training-free inference-time method that edits Whisper decoder activations during generation. The method first estimates a low-rank subspace associated with hallucination-prone non-speech behavior from a calibration split. At inference time, it projects intermediate decoder hidden states away from this subspace, removing the activation component aligned with the estimated directions. To avoid perturbing clear speech inputs, the projection is applied only when Whisper's default no-speech probability exceeds a gate threshold.
\subsection{Calibration Activations}

Let $\ell$ denote a decoder layer and let $d$ denote the hidden-state dimension at that layer. We use a non-speech calibration set to collect decoder hidden representations from two groups of inputs. \emph{The first group contains non-speech examples for which Whisper produces a non-empty hallucinated transcript. The second group contains non-speech examples that are correctly rejected and produce no transcript.}
We denote these layer-$\ell$ hidden representations as
\[
H_\ell
=
\{h^{\mathrm{hall}}_{i,\ell}\}_{i=1}^{N_h},
\qquad
F_\ell
=
\{h^{\mathrm{empty}}_{j,\ell}\}_{j=1}^{N_f},
\]
where $H_\ell$ is the set of hallucinating non-speech representations, $F_\ell$ is the set of correctly empty non-speech representations, $N_h$ and $N_f$ are the numbers of examples in the two groups, and each representation satisfies
$h^{\mathrm{hall}}_{i,\ell},
h^{\mathrm{empty}}_{j,\ell}
\in \mathbb{R}^{d}.
$

The hallucination-associated subspace is estimated only from the calibration set and then kept fixed.
\subsection{Low-Rank Subspace Estimation}

The goal of subspace estimation is to identify decoder directions that distinguish non-speech inputs that trigger hallucinated transcripts from non-speech inputs that are correctly rejected. Since both groups contain non-speech audio, subtracting their activations helps reduce acoustic factors shared by non-speech inputs and emphasizes directions associated with hallucination-prone decoding behavior.

For each decoder layer $\ell$, we construct activation differences between hallucinating and correctly empty non-speech examples:
\[
\Delta_\ell =
\begin{bmatrix}
(h^{\mathrm{hall}}_{1,\ell} - h^{\mathrm{empty}}_{1,\ell})^\top \\
(h^{\mathrm{hall}}_{2,\ell} - h^{\mathrm{empty}}_{2,\ell})^\top \\
\vdots \\
(h^{\mathrm{hall}}_{n,\ell} - h^{\mathrm{empty}}_{n,\ell})^\top
\end{bmatrix}
\in
\mathbb{R}^{n \times d},
\]
where $n=\min(N_h,N_f)$.
We compute the singular value decomposition of the difference matrix 
$
\Delta_\ell
=
U_\ell \Sigma_\ell V_\ell^\top.
$
The right singular vectors corresponding to the largest singular values capture the dominant directions along which hallucinating and correctly empty non-speech examples differ in decoder representation space. We use the top $r$ right singular vectors to define a row-orthonormal projection basis: $
B_{\ell,r}
=
V_{\ell,1:r}^\top
\in
\mathbb{R}^{r \times d}.
$
The rank $r$ controls the dimensionality of the decoder subspace removed at inference time.

\subsection{Inference-Time Decoder Projection}

During inference, we intervene on decoder hidden states at a selected decoder layer. Given a hidden state $h_\ell \in \mathbb{R}^{d}$ and projection basis $B_{\ell,r}$, we replace the hidden state with
\begin{equation}
\tilde{h}_\ell
=
h_\ell
-
\alpha
(h_\ell B_{\ell,r}^{\top})B_{\ell,r},
\label{eq:decoder_projection}
\end{equation}
where $\alpha$ controls the projection strength.

This operation removes the component of the decoder state that lies in the hallucination-associated subspace. If this subspace captures directions that make non-speech inputs more likely to produce text, projecting away from it should reduce hallucinated transcripts while preserving information needed for normal transcription. We intervene on decoder hidden states because Whisper uses the decoder for both transcript generation and no-speech prediction, allowing the method to act before tokens are emitted without changing model parameters or the decoding procedure.
\subsection{No-Speech-Conditioned Gating}

Applying the projection in Eq.~(\ref{eq:decoder_projection}) to every input can perturb real speech. We therefore apply it conditionally using Whisper's own no-speech estimate. Whisper predicts a special \texttt{<|nospeech|>} token for segments without speech~\citep{radford2023robust}; in the official implementation, the decoder probability assigned to this token is exposed as \texttt{no\_speech\_prob} and used for non-speech rejection~\citep{openai_whisper_code}.

Let $p_{\mathrm{ns}}^{\mathrm{base}}$ be the no-speech probability from an initial unprojected Whisper pass. Given a gate threshold $\gamma$, we apply projection only when
$
p_{\mathrm{ns}}^{\mathrm{base}} \geq \gamma .
$
The decoder hidden state is updated as
\[
h'_\ell
=
\begin{cases}
h_\ell - \alpha (h_\ell B_{\ell,r}^{\top})B_{\ell,r},
&
p_{\mathrm{ns}}^{\mathrm{base}} \geq \gamma,
\\[6pt]
h_\ell,
&
p_{\mathrm{ns}}^{\mathrm{base}} < \gamma .
\end{cases}
\]

If the gate is active, Whisper is decoded again with the projection applied; otherwise, the unprojected output is retained. The resulting no-speech probability $p_{\mathrm{ns}}$ is then compared against Whisper's no-speech threshold $\tau$ for the final rejection decision. Thus, $\gamma$ determines when the activation intervention is applied, while $\tau$ determines whether the segment is rejected as no-speech. The full inference procedure is summarized in Algorithm~\ref{alg:gated_projection} in Appendix~\ref{app:algorithm}.
\section{Experimental Setup}
We use the official OpenAI Whisper implementation with the pretrained small, medium, and large-v3 models, and run experiments on an NVIDIA DGX A100 system.
\label{sec:setup}
\subsection{Datasets}
\label{sec:data}
We evaluate our method on both non-speech and speech benchmarks. For non-speech audio, we use ESC-50~\citep{piczak2015esc50}, which contains 2,000 environmental audio clips.  We also use UrbanSound8K~\citep{salamon2014dataset}, which contains 8,732 clips of urban audio clips. Finally, we use FSD50K~\citep{fonseca2021fsd50k}, a large-scale collection of human-labeled sound events. To construct a non-speech subset of FSD50K, we remove clips labeled as speech, vocal, or music content, yielding 8,621 samples.

ESC-50 folds 1--3, containing 1,200 clips, are used for subspace construction and parameter selection. Testing is conducted on held-out ESC-50 folds 4--5, containing 800 clips, as well as UrbanSound8K and the filtered FSD50K subset. Representative examples of Whisper hallucinated transcripts on non-speech audio are provided in Appendix~\ref{app:hallucination_examples}.
To measure whether the method preserves normal speech recognition behavior, we evaluate on LibriSpeech~\citep{panayotov2015librispeech}, a standard read-speech ASR benchmark derived from English audiobook recordings. Unlike the non-speech datasets, LibriSpeech contains spoken samples with reference transcripts, allowing us to measure transcription quality and false rejection of speech as no-speech. We use LibriSpeech validation-clean, containing 2,703 samples, for parameter selection, and reserve test-clean and test-other, containing 2,620 and 2,939 samples, respectively, for final evaluation.




\subsection{Evaluation Metrics and Parameter Selection}
We use Whisper's default no-speech threshold $\tau=0.6$ throughout all experiments. In the gated variant of our method, $\gamma$ denotes the projection gate. Projection is applied only when Whisper's predicted no-speech probability exceeds $\gamma$.

For the non-speech benchmarks, the audio inputs contain no spoken content and therefore have empty reference transcripts. We define a hallucinated transcript as any non-empty output remaining after no-speech filtering and report the HR.
For LibriSpeech, we evaluate speech recognition quality using WER. Since increasing no-speech confidence may incorrectly suppress valid speech, we also report the speech FRR, defined as the fraction of speech samples rejected by the no-speech filter.

We evaluate the proposed method across multiple Whisper model scales. Projection is applied during decoding through a forward hook attached to a selected decoder layer. The method is controlled by four parameters: the decoder layer $\ell$, the projection rank $r$, the projection strength $\alpha$, and the gate threshold $\gamma$.
Because Whisper variants differ in architecture depth and hidden dimensionality, parameter selection is performed separately for each model scale. For each Whisper scale, the final operating parameters $(\ell,r,\alpha,\gamma)$ are  selected once using ESC-50 folds 1--3 together with LibriSpeech validation-clean, where the latter is used only to control degradation on valid speech. 
The resulting configuration is then fixed for all evaluations on the held-out benchmarks described in Section~\ref{sec:data}.

\subsection{Evaluation Baselines}

We compare the proposed method against both internal ablations and existing approaches for mitigating Whisper hallucinations.

\textbf{Original Whisper} denotes the original Whisper model without decoder intervention. When no-speech filtering is enabled, decoding follows Whisper's default rejection mechanism using threshold $\tau$.

\textbf{Always-on projection} applies the learned low-rank projection to all input samples during decoding.

\textbf{Gated projection} first checks whether an input is likely to be non-speech according to Whisper's own no-speech estimate. If this estimate exceeds the gate value $\gamma$, the projection is applied during decoding. Otherwise, the model is left unchanged. This setting is designed to apply the intervention mainly to ambiguous or likely non-speech inputs while avoiding unnecessary changes to clear speech samples.

\textbf{External VAD} filters input audio before transcription using an independent voice activity detector~\citep{SileroVAD}. This baseline represents a standard pipeline-level approach for preventing non-speech segments from reaching the ASR model.

\textbf{Post-hoc hallucination filtering} removes common hallucinated phrases after transcription, such as the BoH approach of \citet{baranski2025investigation}. This baseline operates on generated text rather than model representations.

\textbf{No-speech threshold tuning} modifies the threshold $\tau$ used by Whisper to decide whether audio contains speech. During decoding, the output is discarded when the no-speech probability exceeds $\tau$.

\section{Results and Analysis}
\label{sec:results-analysis}
\subsection{Offline Parameter Selection}
\label{sec:offline-parameter-selection}
\begin{figure*}[t]
    \centering
    \includegraphics[width=0.7\textwidth]{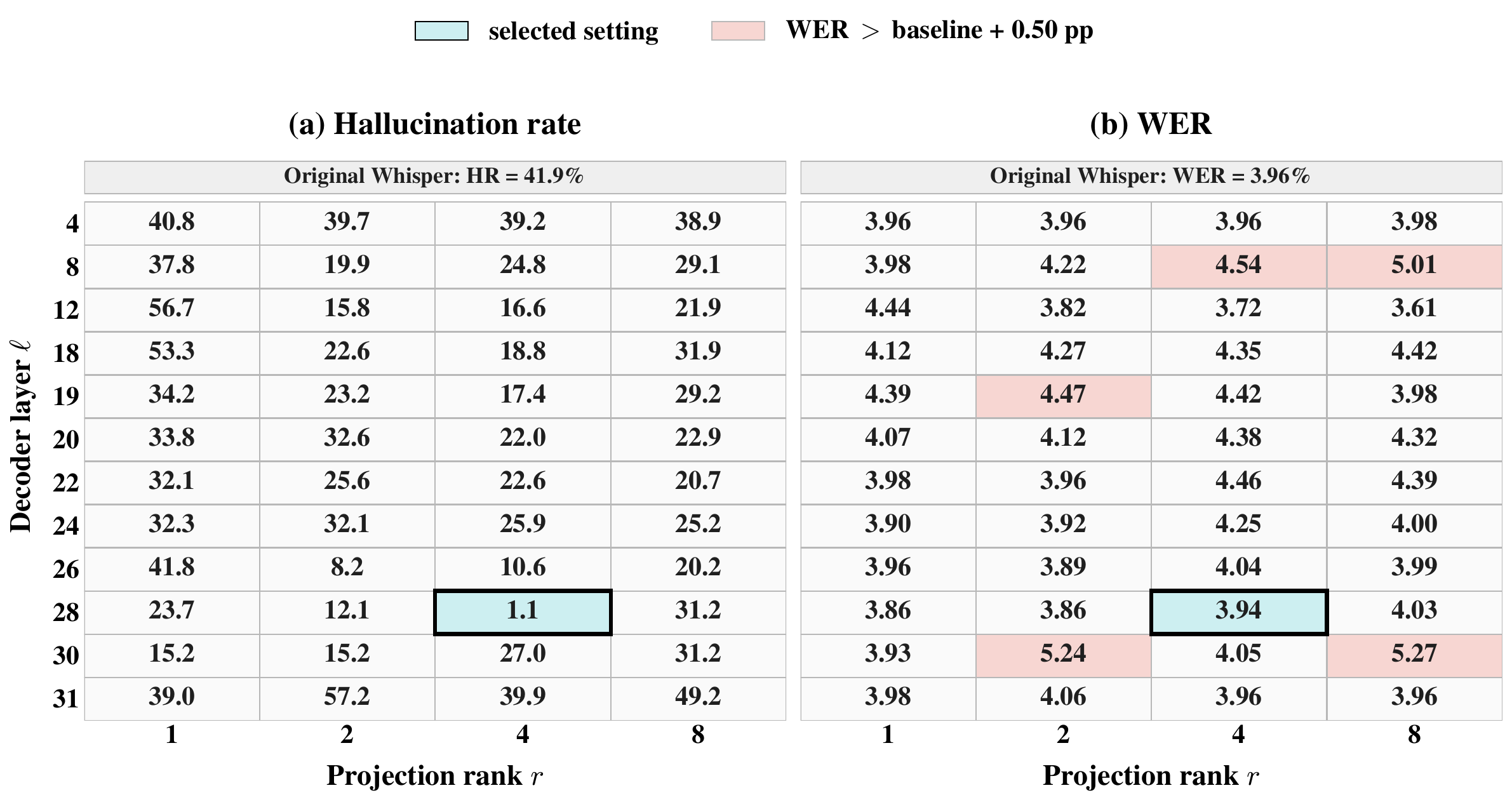}
  \caption{Offline selection of $\ell$ and $r$ for Whisper large-v3. Cells report ESC-50 HR\% and LibriSpeech validation WER\%; lower is better.}
    \label{fig:offline-layer-rank}
\end{figure*}

We determine the projection configuration entirely on the development split and keep it fixed for all held-out evaluations. The search proceeds in two stages. We first choose the decoder layer $\ell$ and projection rank $r$ with $\alpha=1$ and no gating, which allows us to evaluate the intervention location and subspace dimensionality independently of gated activation. Figure~\ref{fig:offline-layer-rank} reports development-set HR on ESC-50 folds 1--3 together with WER on LibriSpeech validation-clean.

The results show that projection effectiveness is primarily determined by decoder depth. Early-layer interventions have little effect on HR, whereas middle-to-late layers substantially reduce hallucinated transcripts. This gain is not without cost, since overly strong interventions can reject valid speech and increase both WER and FRR.

Varying the projection rank shows saturation rather than steady improvement. Small ranks already capture the main hallucination directions, and increasing $r$ does not consistently lower HR. In some settings, larger ranks instead worsen WER and FRR, suggesting unnecessary removal of speech-relevant information.

\begin{figure}[t]
    \centering
    \includegraphics[width=0.7\columnwidth]{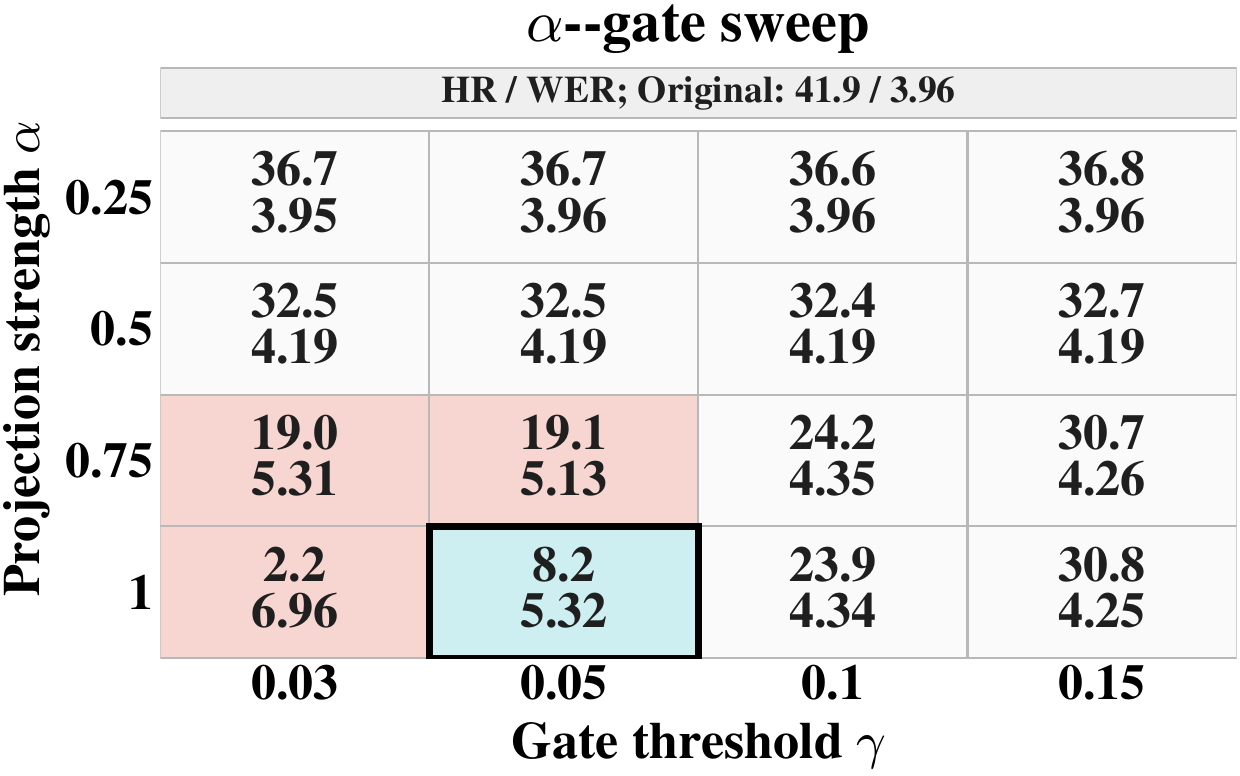}
   \caption{
Offline selection of $\alpha$ and $\gamma$ with $\ell=28$ and $r=4$. Cells show ESC-50 (folds1--3) HR\% and LibriSpeech validation subset WER\%.
}
    \label{fig:offline-alpha-gate}
\end{figure}

For Whisper large-v3, we select the compact setting $\ell=28$ and $r=4$, which gives the best development-set trade-off between hallucination reduction and LibriSpeech performance. We then fix this layer and rank and tune the gated projection parameters, with Figure~\ref{fig:offline-alpha-gate} reporting the resulting development-set HR and WER.

We select $\alpha=1.0$ and $\gamma=0.05$ for all subsequent experiments. On the development set, this configuration reduces HR from 41.92\% to 8.2\%. Although \emph{always-on} projection can achieve lower HR, it does so by rejecting many speech samples. On LibriSpeech validation-clean, always-on projection increases WER to 11.59\% with 9.69\% FRR, whereas gated projection achieves 5.32\% WER with 1.85\% FRR. \emph{We therefore use gated projection as the final setting, since it retains strong hallucination suppression while substantially reducing speech false rejection.}
For Whisper small and medium models, the selected settings are $\ell=10$, $r=1$, $\alpha=1.0$, $\gamma=0.15$ and $\ell=24$, $r=2$, $\alpha=0.75$, $\gamma=0.10$, respectively. Detailed offline selection tables for these models are provided in Appendix~\ref{app:offline-selection-other-scales}.

\subsection{Hallucination Reduction and Cross-Dataset Generalization}
\label{sec:hallucination-generalization}

Table~\ref{tab:hallucination-generalization} reports HR on non-speech audio using Whisper's default no-speech filtering threshold $\tau=0.6$. We compare original Whisper, \emph{always-on} projection, and \emph{gated} projection.

For Whisper large-v3, original Whisper still produces non-empty transcripts for 44.25\% of held-out ESC-50 examples. Always-on projection achieves the strongest reduction, lowering hallucination rate to 1.50\%, while gated projection reduces it to 8.38\%. These results show that decoder projection substantially improves rejection of non-speech inputs.
Our method also generalizes strongly beyond the calibration dataset. On UrbanSound8K, hallucination rate decreases from 76.08\% with original Whisper to 0.87\% with always-on projection and 2.74\% with gated projection. On the filtered FSD50K subset, it decreases from 21.35\% to 0.18\% and 1.15\%, respectively. Similar trends are observed across Whisper small and medium models (Table~\ref{tab:hallucination-generalization}). \emph{This cross-dataset transfer suggests that the projection targets a reusable decoder-level signature of non-speech hallucination, rather than acoustic properties specific to ESC-50.}

\begin{table}[t]
\centering
\footnotesize
\setlength{\tabcolsep}{3pt}
\begin{tabular}{llccc}
\hline
\textbf{Model} & \textbf{Method} & \textbf{ESC-50} & \textbf{Urban8K} & \textbf{FSD50K} \\
\hline
Small
& Original & 23.50 & 12.33 & 21.35 \\
& Always-on & 1.50 & 0.81 & 6.68 \\
& Gated & 1.53 & 0.84 & 6.84 \\
\hline
Medium
& Original & 26.50 & 14.52 & 41.95 \\
& Always-on & 1.82 & 0.62 & 8.02 \\
& Gated & 2.75 & 0.66 & 8.78 \\
\hline
Large-v3
& Original & 44.25 & 76.08 & 21.35 \\
& Always-on & 1.50 & 0.87 & 0.18 \\
& Gated & 8.38 & 2.74 & 1.15 \\
\hline
\end{tabular}
\caption{Non-speech hallucination rate (HR, \%) for Whisper models on ESC-50
(folds 4--5), UrbanSound8K, and FSD50K, with Whisper's
no-speech threshold set to $\tau=0.6$.}
\label{tab:hallucination-generalization}
\end{table}

\subsection{Speech Recognition and False Rejection}
\label{sec:speech-recognition-false-rejection}
Hallucination mitigation should not substantially degrade normal speech recognition or incorrectly reject valid speech as non-speech. We therefore evaluate LibriSpeech under the same three conditions used in the non-speech experiments. For each setting, we report WER as a measure of transcription quality together with speech FRR, which measures the fraction of speech samples incorrectly filtered as no-speech.

Table~\ref{tab:librispeech-wer} reports LibriSpeech WER across Whisper model scales. For Whisper large-v3, original Whisper obtains 4.06\% WER on test-clean and 5.87\% on test-other. Gated projection obtains 6.17\% and 6.57\%, respectively, while always-on projection increases WER to 12.95\% and 13.13\%. \emph{This shows that the gating projection substantially reduces the speech degradation caused by applying projection to all inputs.}

\begin{table}[t]
\centering
\footnotesize
\setlength{\tabcolsep}{3pt}
\begin{tabular}{llcc}
\hline
\textbf{Model} & \textbf{Method} & \textbf{Clean WER} & \textbf{Other WER} \\
\hline
Small
& Original & 4.04 & 8.38 \\
& Always-on & 4.51 & 11.48 \\
& Gated & 4.37 & 9.13 \\
\hline
Medium
& Original & 3.66 & 7.29 \\
& Always-on & 6.40 & 15.06 \\
& Gated & 5.47 & 11.68 \\
\hline
Large-v3
& Original & 4.06 & 5.87 \\
& Always-on & 12.95 & 13.13 \\
& Gated & 6.17 & 6.57 \\
\hline
\end{tabular}
\caption{LibriSpeech word error rate (WER, \%) for Whisper models on test-clean and test-other. Lower is better.}
\label{tab:librispeech-wer}
\end{table}

Table~\ref{tab:librispeech-false-rejection} reports FRR. For Whisper large-v3, original Whisper almost never rejects LibriSpeech speech samples as no-speech. Always projection is substantially more aggressive, falsely rejecting 10.50\% of test-clean and 11.47\% of test-other samples. Gated projection substantially reduces this failure mode, lowering FRR to 2.86\% on test-clean and 1.40\% on test-other while preserving strong hallucination suppression on non-speech audio.

\begin{table}[t]
\centering
\footnotesize
\setlength{\tabcolsep}{3pt}
\begin{tabular}{llcc}
\hline
\textbf{Model} & \textbf{Method} & \textbf{Clean FRR} & \textbf{Other FRR} \\
\hline
Small
& Original & 0.00 & 0.00 \\
& Always-on & 0.64 & 4.86 \\
& Gated & \textbf{0.41} & \textbf{2.58} \\
\hline
Medium
& Original & 0.03 & 0.27 \\
& Always-on & 5.68 & 16.87 \\
& Gated & \textbf{4.07} & \textbf{9.97} \\
\hline
Large-v3
& Original & 0.04 & 0.00 \\
& Always-on & 10.50 & 11.47 \\
& Gated & \textbf{2.86} & \textbf{1.40} \\
\hline
\end{tabular}
\caption{LibriSpeech false rejection rate (FRR, \%) for Whisper models on test-clean and test-other at $\tau=0.6$. Lower is better.}
\label{tab:librispeech-false-rejection}
\end{table}

\begin{figure*}[t]
    \centering
    \includegraphics[width=0.8\textwidth]{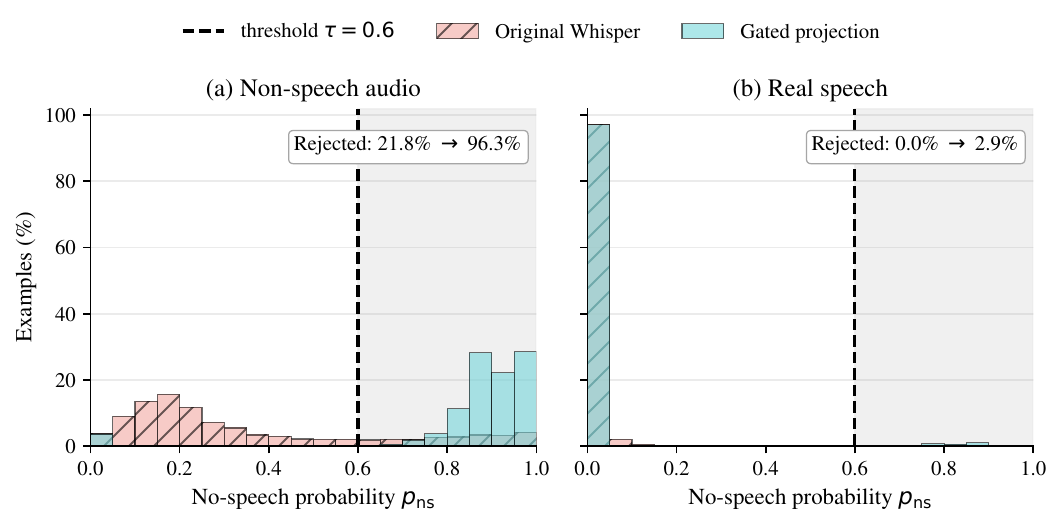}
\caption{No-speech probability before and after gated projection on ESC-50 and LibriSpeech test-clean; dashed line marks $\tau=0.6$.}
    \label{fig:nospeech-before-after}
\end{figure*}

To better understand this behavior, Figure~\ref{fig:nospeech-before-after} compares Whisper's no-speech probability before and after gated projection. On ESC-50, projection shifts many non-speech examples above the rejection threshold, making them more likely to be filtered as no-speech. On LibriSpeech test-clean, the distribution remains concentrated near zero, indicating that the intervention has a limited effect on clear speech samples. These observations are consistent with the aggregate results in Tables~\ref{tab:hallucination-generalization} and~\ref{tab:librispeech-false-rejection}, where projection substantially reduces hallucinated transcripts on non-speech audio while  preserving normal speech recognition behavior.These results highlight the intended deployment trade-off. Gated projection is the preferred deployment mode because it retains strong hallucination suppression while substantially reducing the false rejection introduced by always-on projection. Importantly, false rejection need not imply permanent loss of speech: uncertain segments can be retained and flagged for re-transcription, verification, or human review rather than being discarded. This is particularly useful in reliability-sensitive settings where unsupported transcripts may lead to consequential downstream decisions or actions. Overall, the method provides strong hallucination control while remaining training-free and requiring neither an external model nor modifications to Whisper’s parameters.

We include additional ablations in Appendix~\ref{app:ablation}. We first vary Whisper's no-speech threshold $\tau$ after gated projection and find that the default $\tau=0.6$ gives the best trade-off. We also test projection at two decoder layers, which further reduces non-speech hallucinations but substantially increases LibriSpeech WER and speech FRR. \textit{These results support using single-layer gated projection with the default no-speech threshold.}
\subsection{Comparison with Baselines}
\label{sec:baseline-comparison}
\begin{figure*}[!h]
    \centering    \includegraphics[width=\textwidth]{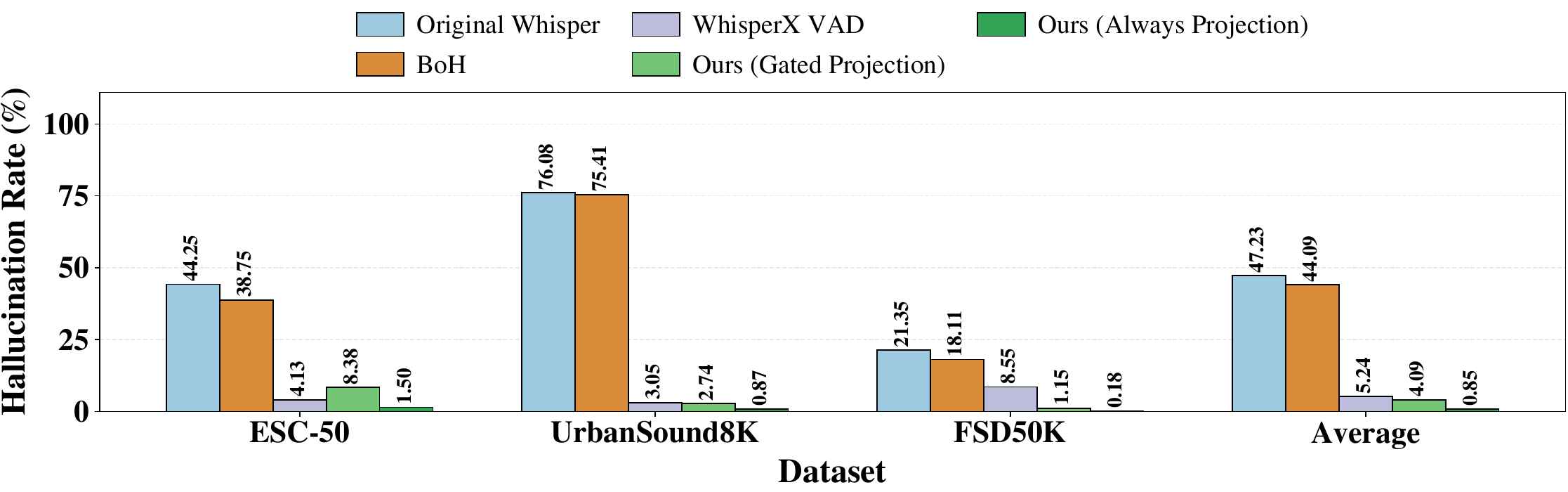}
    \caption{
    Baseline comparison on non-speech datasets. Bars show hallucination rate (HR, \%); lower is better. 
    }
    \label{fig:baseline-comparison}
\end{figure*}
We compare against training-free baselines that mitigate Whisper hallucinations through input filtering, output filtering, and no-speech threshold tuning. For output filtering, we reimplement the BoH approach of \citet{baranski2025investigation} because no public implementation is available. We build a list of frequent hallucinated phrases from the ESC-50 development split and remove matching phrases after decoding; the phrase list is provided in Appendix~\ref{app:boh_phrases}. For input filtering, we use WhisperX VAD~\citep{bain2023whisperx}, which removes regions predicted as non-speech before transcription. 
Figure~\ref{fig:baseline-comparison} reports hallucination rates on non-speech datasets. Original Whisper produces high hallucination rates, with 44.25\% on ESC-50, 76.08\% on UrbanSound8K, and 21.35\% on FSD50K. The BoH implementation reduces some recurring hallucinated phrases, but remains limited, with HRs of 38.75\%, 75.41\%, and 18.11\%. This is consistent with the behavior reported by \citet{baranski2025investigation}, where phrase-list filtering can suppress repeated patterns such as common acknowledgments or subtitle-like endings, but cannot remove hallucinations that do not overlap with the calibration list.

WhisperX VAD is a stronger training-free baseline, reducing average HR to 5.24\% across the three datasets. Our Gated projection obtains a lower average HR of 4.09\%, while always-on projection performs best with an average HR of 0.85\%.

We also report speech recognition performance to make the HR--WER/FRR trade-off explicit. On LibriSpeech, with threshold tuning at $\tau=0.5$, Whisper achieves WERs of 4.15\% on test-clean and 5.87\% on test-other, with FRRs of 0.08\% and 0\%, respectively, but only modestly reduces the hallucination rate to 45.99\% (Appendix~\ref{app:threshold-sweep}).
WhisperX achieves lower WERs of 2.67\% and 4.74\%, with 0\% FRR on both sets, while reducing average HR to 5.24\%; however, it requires an external VAD and segmentation stage. Our gated projection further reduces average HR to 4.09\% without an external detector, with WERs of 6.17\% and 6.57\% and FRRs of 2.86\% and 1.40\% on test-clean and test-other, respectively. Always-on projection achieves the lowest average HR of 0.85\%, but at a larger cost in WER and FRR. Overall, gated projection offers a favorable trade-off between hallucination suppression and speech preservation.

For comparison with model fine-tuning-based mitigation, we include the reported results of Calm-Whisper(Wangetal.,2025). Calm-Whisper~\citep{wang2025calm} reports 15.51\% HR on UrbanSound8K and LibriSpeech WERs of 2.19\% on test-clean and 4.13\% on test-other. Because no public checkpoint is available, we do not re-evaluate it in our pipeline. On UrbanSound8K, our gated and always-on variants achieve HRs of 2.74\% and 0.87\%, respectively, without updating Whisper parameters.
Overall, gated projection provides a controllable HR--WER/FRR trade-off: projection modifies the decoder representation, while the no-speech threshold $\tau$ selects the final rejection point. Unlike threshold tuning, projection acts before the no-speech decision by suppressing a hallucination-associated decoder subspace, without external VAD, phrase lists, fine-tuning, or parameter updates.
\section{Conclusion}
We introduced a training-free decoder projection method for reducing Whisper hallucinations on non-speech audio. The method estimates a compact hallucination-associated subspace and edits decoder representations at inference time without parameter updates. Across multiple datasets and Whisper scales, it substantially reduces hallucinated transcripts while providing a controllable HR--WER--FRR trade-off. These results indicate that hallucination-associated decoder directions can be manipulated through lightweight inference-time intervention.
\section*{Limitations} Our study focuses on mitigating non-speech hallucinations in Whisper through a training-free decoder-level intervention. This scope reflects a common and practically important failure mode in which non-speech audio is transcribed as fluent text. However, hallucinations in ASR can also arise in other settings, such as long-form transcription, multilingual audio, or acoustically ambiguous inputs. Our method is designed and evaluated for the non-speech setting, and extending the analysis to these broader hallucination scenarios remains an important direction for future work. 
The method introduces an HR--WER--FRR trade-off. Gated projection reduces the speech degradation of always-on projection, but some operating points still increase WER and FRR. In applications where missed speech is costly, less aggressive settings may be preferred, and uncertain segments can be retained and flagged for re-transcription or verification rather than permanently discarded.

Finally, the subspace is estimated from non-speech calibration data. Although it transfers unchanged from ESC-50 to held-out ESC-50, UrbanSound8K, and FSD50K, broader calibration-to-deployment shifts remain to be evaluated. The gated variant also depends partly on the reliability of Whisper's internal no-speech probability.

\section*{Ethical Considerations}

We do not collect new audio, annotations, or human-subject data. Our experiments use the official OpenAI Whisper implementation and pretrained Whisper model weights, which are released under the MIT License, together with public benchmark datasets including ESC-50, UrbanSound8K, FSD50K, and LibriSpeech. We use these artifacts only for research evaluation of ASR hallucination behavior and do not redistribute the datasets, release derived audio, or use derivatives outside the research context.

We document the artifacts used in our experiments by specifying their domains and roles. ESC-50 and UrbanSound8K are used as environmental and urban non-speech audio benchmarks, FSD50K is used through a filtered non-speech subset, and LibriSpeech is used as an English read-speech ASR benchmark. We report the number of samples and the development/test splits used for each dataset in the experimental setup.

LibriSpeech contains human speech recordings. We report only aggregate ASR metrics and do not attempt speaker identification, demographic inference, or release speaker-level information. For FSD50K, we remove clips labeled as speech, vocal, or music content when constructing the non-speech subset.

No new human annotators or participants are recruited for this study. The method modifies inference-time decoder activations of an existing ASR model and does not update Whisper's parameters.

\bibliography{custom}
\appendix
\section{Gated Low-Rank Decoder Projection Algorithm}
\label{app:algorithm}

Algorithm~\ref{alg:gated_projection} summarizes the inference-time procedure used for gated low-rank decoder projection. The method first uses Whisper's default no-speech probability to decide whether an input is likely to be non-speech. Projection is only activated when this probability exceeds the gate threshold $\gamma$; otherwise, the model decodes normally. This avoids applying the intervention to most speech inputs.
\begin{algorithm}[t]
\caption{Gated Low-Rank Decoder Projection}
\label{alg:gated_projection}
\begin{algorithmic}[1]
\Require Whisper model $M$, audio $x$, decoder layer $\ell$, basis $B_{\ell,r}$, projection strength $\alpha$, gate threshold $\gamma$, rejection threshold $\tau$
\State Run Whisper without projection to obtain unprojected transcript $y^{\mathrm{base}}$ and unprojected no-speech probability $p_{\mathrm{ns}}^{\mathrm{base}}$
\If{$p_{\mathrm{ns}}^{\mathrm{base}} \geq \gamma$}
    \State Attach projection hook at decoder layer $\ell$
    \State During decoding, replace hidden states by
    \[
    h'_\ell = h_\ell - \alpha(h_\ell B_{\ell,r}^{\top})B_{\ell,r}
    \]
    \State Decode again to obtain transcript $y$ and no-speech probability $p_{\mathrm{ns}}$
\Else
    \State Set $y \leftarrow y^{\mathrm{base}}$ and $p_{\mathrm{ns}} \leftarrow p_{\mathrm{ns}}^{\mathrm{base}}$
\EndIf
\If{$p_{\mathrm{ns}} > \tau$}
    \State Return empty transcript
\Else
    \State Return $y$
\EndIf
\end{algorithmic}
\end{algorithm}
Here, $B_{\ell,r}$ denotes the rank-$r$ hallucination-associated basis estimated from calibration activations at decoder layer $\ell$. The gate threshold $\gamma$ controls whether projection is applied, while the rejection threshold $\tau$ controls Whisper's final no-speech filtering decision. In our main experiments, the basis is estimated once and then kept fixed across all evaluation datasets.
\section{Examples of Whisper Hallucinations on Non-Speech Audio}
\label{app:hallucination_examples}
Representative examples of hallucinated transcripts produced by Whisper on non-speech audio are shown in Table~\ref{tab:hallucination_examples}. These examples illustrate that non-speech inputs can trigger fluent but unrelated textual continuations despite the absence of spoken content.
\begin{table}[t]
\centering
\small
\setlength{\tabcolsep}{3pt}
\renewcommand{\arraystretch}{1.12}
\fcolorbox{black!25}{black!3}{
\begin{minipage}{0.96\columnwidth}
\begin{tabularx}{\linewidth}{@{}p{0.28\linewidth}p{0.24\linewidth}Y@{}}
\toprule
Source & Audio event & Whisper transcript \\
\midrule
ESC-50
& rooster
& ``i'm the best'' \\
\midrule
ESC-50
& car horn
& ``the train is coming up'' \\
\midrule
UrbanSound8K
& jackhammer
& ``thank you'' \\
\midrule
UrbanSound8K 
& engine idling
& "so" \\
\bottomrule

FSD50K 
& dog barking
& "dog, dog, dog, dog, dog" \\
\bottomrule

FSD50K 
&Frying
& "you" \\
\bottomrule
\end{tabularx}

\end{minipage}
}
\caption{Representative hallucinated transcripts produced by Whisper on non-speech audio.All audio clips are from non-speech benchmarks and contain no intended spoken transcript. Any non-empty transcript is therefore treated as a hallucinated transcription.}
\label{tab:hallucination_examples}
\end{table}

\section{Additional Results for Whisper Small and Medium}
\label{app:offline-selection-other-scales}

\begin{figure*}[t]
    \centering
    \includegraphics[width=0.7\textwidth]{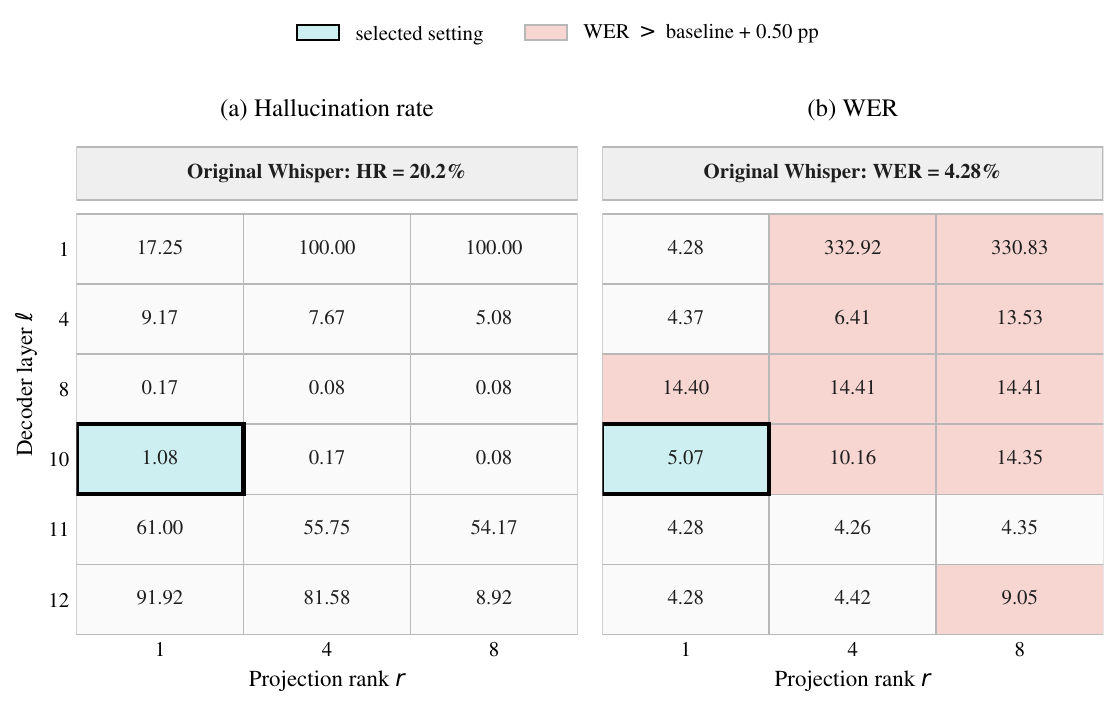}
    \caption{
    Offline selection of decoder layer $\ell$ and projection rank $r$ for Whisper small using always-on projection with fixed $\alpha=1$. The left panel reports development-set hallucination rate (HR\%) on ESC-50 folds 1--3, and the right panel reports WER \% on LibriSpeech validation-clean subset.
    }
    \label{fig:offline-small}
\end{figure*}

\begin{figure*}[t]
    \centering
    \includegraphics[width=0.7\textwidth]{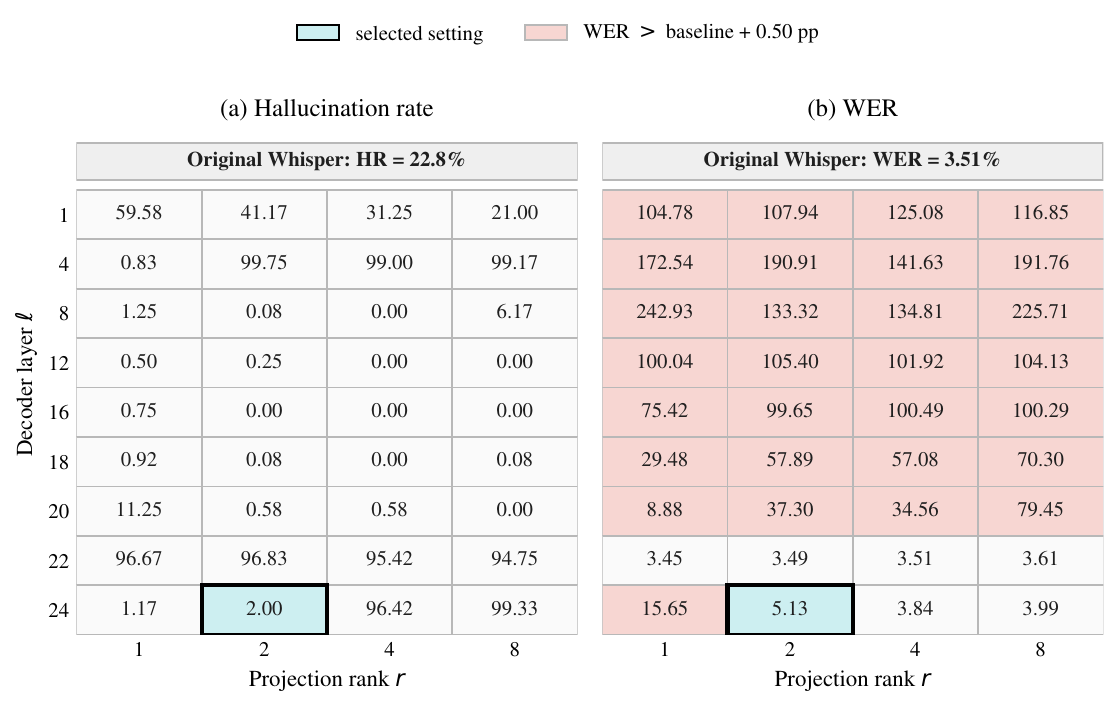}
    \caption{
    Offline selection of decoder layer $\ell$ and projection rank $r$ for Whisper medium using always-on projection with fixed $\alpha=1$. The left panel reports development-set hallucination rate (HR\%) on ESC-50 folds 1--3, and the right panel reports WER\% on LibriSpeech validation-clean.
    }
    \label{fig:offline-medium}
\end{figure*}

Figures~\ref{fig:offline-small} and~\ref{fig:offline-medium} report the offline parameter analysis for Whisper small and medium models. As with Whisper large-v3, projection effectiveness depends strongly on decoder layer and projection rank. Early-layer projection yields limited hallucination reduction, whereas middle-to-late layers achieve substantially lower hallucination rates.

\begin{figure}[t]
    \centering
    \includegraphics[width=0.85\columnwidth]{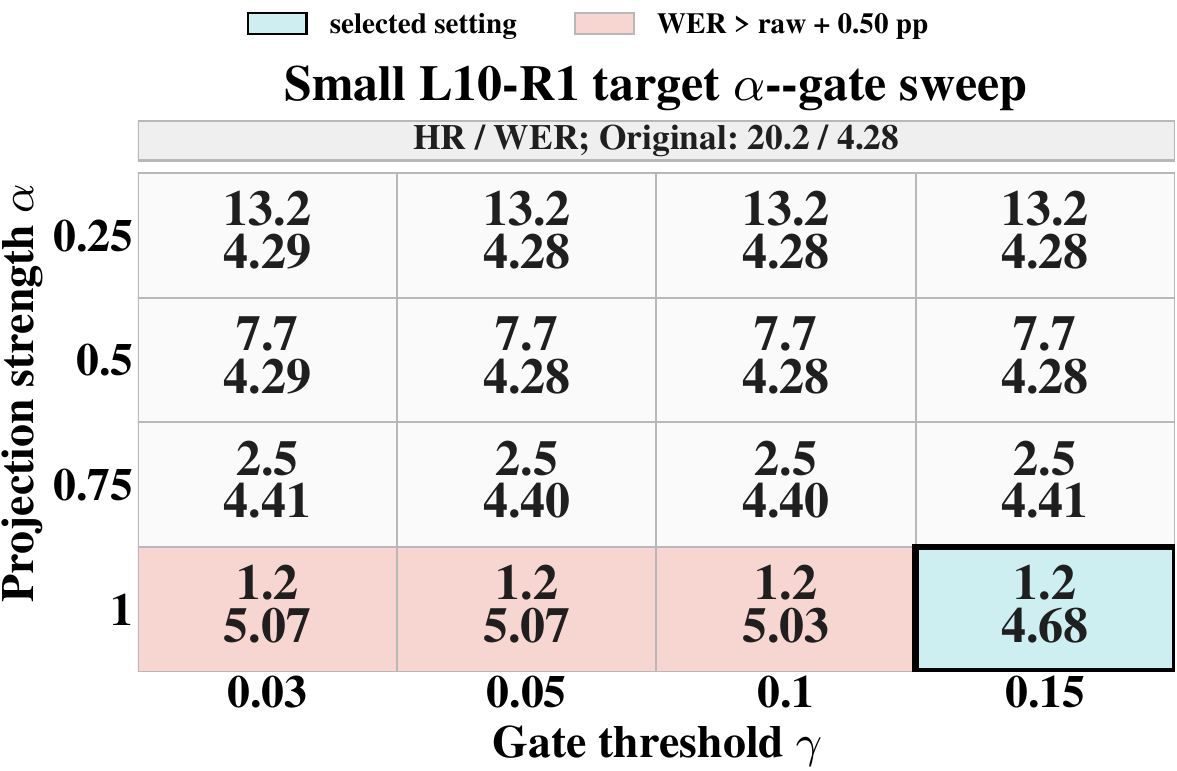}
    \caption{
    Offline sweep of projection strength $\alpha$ and gate threshold $\gamma$ for Whisper small after fixing $\ell=10$ and $r=1$. Cells report development-set HR and WER.
    }
    \label{fig:offline-small-gated}
\end{figure}

For Whisper small, the best non-gated trade-off is obtained at layer $\ell=10$ with rank $r=1$, reducing HR from 20.2\% to 1.08\% while increasing WER from 4.28\% to 5.07\%. Figure~\ref{fig:offline-small-gated} shows the subsequent gated sweep. We select $\alpha=1.0$ and $\gamma=0.15$, which further improves the trade-off by reducing HR to 1.2\% with 4.68\% WER.

\begin{figure}[t]
    \centering
    \includegraphics[width=0.75\columnwidth]{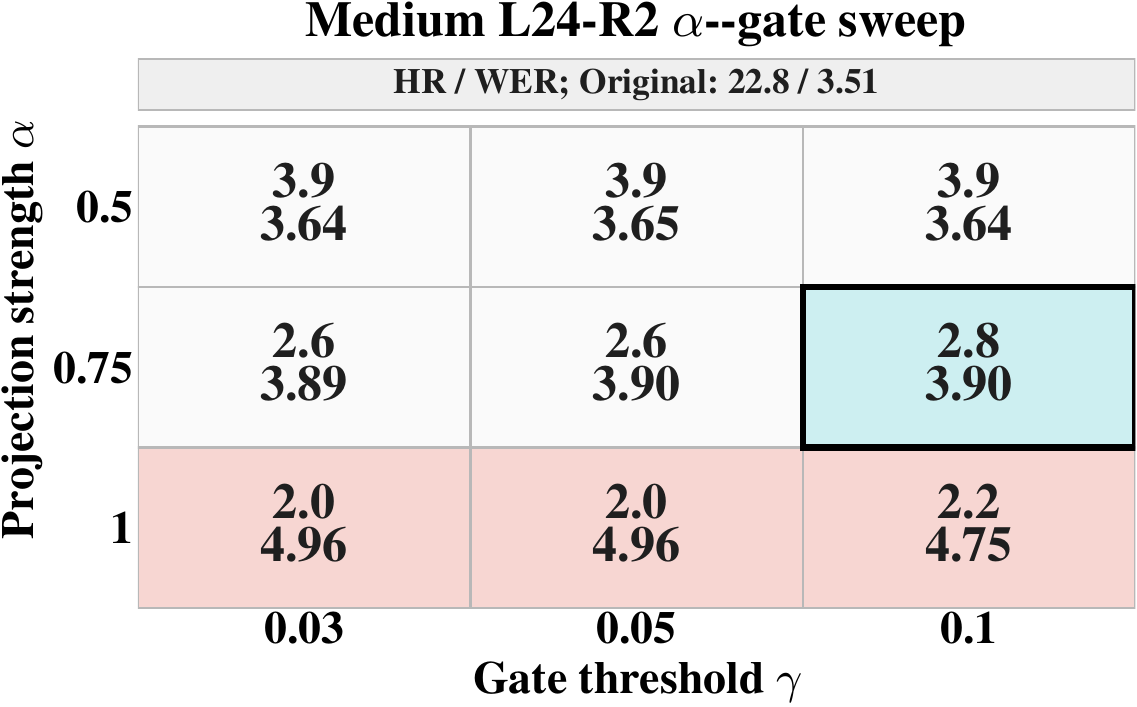}
    \caption{
    Offline sweep of projection strength $\alpha$ and gate threshold $\gamma$ for Whisper medium after fixing $\ell=24$ and $r=2$. Cells report development-set HR\% and WER\%.
    }
    \label{fig:offline-medium-gated}
\end{figure}

For Whisper medium, the best non-gated configuration is obtained at layer $\ell=24$ with rank $r=2$, reducing HR from 22.8\% to 2.0\% with 5.13\% WER. Figure~\ref{fig:offline-medium-gated} shows the gated sweep after fixing the selected layer and rank. We select $\alpha=0.75$ and $\gamma=0.10$, which gives 2.8\% HR with 3.90\% WER. Similar to Whisper large-v3, gated projection provides a substantially better balance between hallucination suppression and speech preservation than always-on projection.

\section{Ablation Study}
\label{app:ablation}
\subsection{ No-Speech Threshold Tuning with Gated Projection}
We examine whether no-speech threshold tuning can further improve our gated projection results. Whisper's default threshold ($\tau=0.6$) determines when an output is discarded as non-speech. Since projection changes Whisper's no-speech probabilities, the default threshold may no longer be the best operating point after projection. This ablation therefore evaluates our gated projection method under different values of $\tau$.

We use Whisper-medium and fix the projection configuration to $\ell=24$, $r=2$, $\alpha=1.0$, and $\gamma=0.10$. We vary the no-speech threshold $\tau$ and, for each value, compare original Whisper with gated projection under the same threshold. We report average HR over ESC-50 (folds 4 ,5), UrbanSound8K, and FSD50K, average WER over LibriSpeech test-clean and test-other, and speech FRR on LibriSpeech.

The results show that threshold tuning changes the operating point of the projected system. Compared with the default projected setting at $\tau=0.6$, lowering the threshold to $\tau=0.4$ further reduces average HR from $5.09\%$ to $2.89\%$, but increases average WER from $8.58\%$ to $11.49\%$ and speech FRR from $7.03\%$ to $10.91\%$. Raising the threshold to $\tau=0.7$ has the opposite effect. It reduces average WER to $7.28\%$ and speech FRR to $4.76\%$, while increasing average HR to $7.85\%$.

Thus, no-speech threshold tuning can be used together with gated projection to choose a desired operating point. The default threshold $\tau=0.6$ provides the best trade-off, while smaller or larger thresholds prioritize lower hallucination rate or lower speech rejection, respectively.
\begin{table}[t]
\centering
\small
\setlength{\tabcolsep}{4pt}
\begin{tabular}{c l c c c}
\hline
$\tau$ & Method & Avg. HR $\downarrow$ & Avg. WER $\downarrow$ & Avg. FRR $\downarrow$ \\
\hline
0.4 & Original & 13.08 & 5.77 & 0.80 \\
    & +Proj.   & 2.89  & 11.49 & 10.91 \\
\hline
0.5 & Original & 20.42 & 5.54 & 0.35 \\
    & +Proj.   & 3.78  & 9.90 & 9.01 \\
\hline
0.6 & Original & 29.49 & 5.48 & 0.14 \\
    & +Proj.   & 5.09  & 8.58 & 7.03 \\
\hline
0.7 & Original & 42.75 & 5.46 & 0.09 \\
    & +Proj.   & 7.85  & 7.28 & 4.76 \\
\hline
\end{tabular}
\caption{
No-speech threshold ablation for Whisper-medium. HR is averaged over non-speech datasets; WER and speech FRR are averaged over LibriSpeech test-clean and test-other subsets.
}
\label{tab:threshold-projection-ablation}
\end{table}
\subsection{Multi-Layer Gated Projection}
\label{sec:ablation-multilayer}

We examine whether applying projection at more than one decoder layer further reduces non-speech hallucinated transcripts. Based on the offline parameter selection for Whisper-medium, we choose two strong intervention layers, $\ell=20$ and $\ell=24$. To avoid excessive degradation in speech recognition accuracy, we use a smaller projection strength ($\alpha$) for the additional layer. The multi-layer setting applies $(\ell=20,r=4,\alpha=0.5)$ and $(\ell=24,r=2,\alpha=1.0)$ with gated projection. We fix the gate threshold to $\gamma=0.10$ and Whisper's no-speech threshold to $\tau=0.6$.

\begin{table*}[t]
\centering
\small
\setlength{\tabcolsep}{5pt}
\begin{tabular}{lccc|cc|cc}
\hline
\multirow{2}{*}{Method|} &
\multicolumn{3}{c|}{HR $\downarrow$} &
\multicolumn{2}{c|}{WER $\downarrow$} &
\multicolumn{2}{c}{Speech FRR $\downarrow$} \\
& ESC-50 & US8K & FSD50K & clean & other & clean & other \\
\hline
Original Whisper
& 26.50 & 14.52 & 41.95 & 3.67 & 7.30 & 0.03 & 0.27 \\

Single-layer proj.
& 2.75 & 0.66 & 8.78 & 5.48 & 11.68 & 4.07 & 9.97 \\

Multi-layer proj.
& 0.13 & 0.07 & 4.87 & 10.62 & 15.71 & 11.07 & 13.75 \\
\hline
\end{tabular}
\caption{
Multi-layer projection ablation for Whisper-medium. HR is reported on ESC-50 (folds 4-5), UrbanSound8K, and FSD50K; WER and speech FRR are reported on LibriSpeech test-clean and test-other subsets.
}
\label{tab:multilayer-ablation}
\end{table*}

Table~\ref{tab:multilayer-ablation} shows that multi-layer projection further reduces hallucination rates on all non-speech datasets. Compared with the single-layer setting, HR decreases from $2.75\%$ to $0.13\%$ on ESC-50, from $0.66\%$ to $0.07\%$ on UrbanSound8K, and from $8.78\%$ to $4.87\%$ on FSD50K. This suggests that hallucination-related directions are distributed across multiple decoder layers, and that combining projections can provide stronger non-speech suppression. However, this improvement comes with increased speech degradation. On LibriSpeech dataset, WER increases from $5.48\%$ to $10.62\%$ on test-clean subset and from $11.68\%$ to $15.71\%$ on test-other subset. Speech FRR also increases from $4.07\%$ to $11.07\%$ on test-clean and from $9.97\%$ to $13.75\%$ on test-other. \textit{We therefore keep single-layer gated projection as the main setting. Multi-layer projection is effective for stronger hallucination suppression, but the single-layer configuration gives a better operating point for preserving speech recognition.}

\section{Bag of Hallucinations Phrase List}
\label{app:boh_phrases}
\begin{table}[t]
\centering
\small
\setlength{\tabcolsep}{3pt}
\renewcommand{\arraystretch}{1.12}
\fcolorbox{black!25}{black!3}{
\begin{minipage}{0.96\columnwidth}
\begin{tabularx}{\linewidth}{@{}Y@{}}
\toprule
\textbf{Frequent hallucinated phrase} \\
\midrule
``oh my god'' \\
``thank you'' \\
``i'm going to go to the right'' \\
``i'm sorry'' \\
``i'm not sure what i'm doing here'' \\
``meow meow'' \\
``the end'' \\
``the train is coming'' \\
``bf watch tv 2021'' \\
''thank you for watching''\\
''you''\\
\bottomrule
\end{tabularx}
\end{minipage}
}
\caption{
Frequent hallucinated phrases used by the BoH filtering baseline.
}
\label{tab:boh_phrases}
\end{table}

Table~\ref{tab:boh_phrases} lists the frequent hallucinated phrases used for the BoH filtering baseline. The phrases are extracted from Whisper outputs on the ESC-50 development split and removed from decoded transcripts when they appear as exact normalized matches.

\section{No-Speech Threshold Sweep}
\label{app:threshold-sweep}

Table~\ref{tab:threshold-sweep} reports a threshold-only baseline for original Whisper, where we vary the no-speech threshold $\tau$ without applying projection. Lowering $\tau$ reduces non-speech hallucinations, with $\tau=0.5$ giving the lowest average HR, but the improvement over the default $\tau=0.6$ is modest. Speech recognition metrics remain largely unchanged across thresholds, indicating that threshold tuning alone is insufficient to substantially mitigate non-speech hallucinations.
\begin{table*}[t]
\centering
\small
\begin{tabular}{lrrrrrrrr}
\hline
$\tau$ & ESC HR & Urban HR & FSD HR & Avg HR & Clean WER & Clean FRR & Other WER & Other FRR \\
\hline
0.5 & 44.00 & 72.55 & 21.43 & 45.99 & 4.15 & 0.08 & 5.87 & 0.00 \\
0.6 & 44.25 & 76.08 & 21.35 & 47.23 & 4.07 & 0.04 & 5.87 & 0.00 \\
0.7 & 48.88 & 79.62 & 40.16 & 56.22 & 4.07 & 0.04 & 5.87 & 0.00 \\
0.8 & 54.37 & 83.76 & 54.26 & 64.13 & 4.03 & 0.00 & 5.87 & 0.00 \\
0.9 & 59.13 & 89.50 & 68.45 & 72.36 & 4.03 & 0.00 & 5.87 & 0.00 \\
\hline
\end{tabular}
\caption{
Threshold-only baseline for original Whisper under different no-speech thresholds $\tau$. HR\% is averaged over ESC-50 (folds 4-5), UrbanSound8K, and FSD50K. WER\% and FRR\% are reported on LibriSpeech test-clean and test-other subsets.
}
\label{tab:threshold-sweep}
\end{table*}

\end{document}